\documentclass[fleqn,10pt]{wlscirep}
\usepackage[utf8]{inputenc}
\usepackage[T1]{fontenc}

\usepackage{bm}
\usepackage{overpic}

\usepackage{color}
\usepackage{ulem}

\DeclareMathOperator*{\argmin}{arg\,min}

\title{Lossy compression of matrices by black box optimisation of mixed integer nonlinear programming}

\author[1,2,*]{Tadashi Kadowaki}
\author[3]{Mitsuru Ambai}
\affil[1]{DENSO CORPORATION, Kounan, Minato-ku, Tokyo 108-0075, Japan}
\affil[2]{Research Center for Emerging Computing Technologies (RCECT),
National Institute of Advanced Industrial Science and Technology (AIST),
Umezono, Tsukuba, Ibaraki 305-8568, Japan}
\affil[3]{DENSO IT Laboratory, Shibuya, Shibuya-ku, Tokyo 150-0002, Japan}

\affil[*]{tadashi.kadowaki.j3m@jp.denso.com}

\begin{abstract}
In edge computing, suppressing data size is a challenge for machine learning models that perform complex tasks such as autonomous driving, in which computational resources (speed, memory size and power) are limited.
Efficient lossy compression of matrix data has been introduced by decomposing it into the product of an integer and real matrices.
However, its optimisation is difficult as it requires simultaneous optimisation of an integer and real variables.
In this paper, we improve this optimisation by utilising recently developed black-box optimisation (BBO) algorithms with an Ising solver for binary variables.
In addition, the algorithm can be used to solve mixed-integer programming problems that are linear and non-linear in terms of real and integer variables, respectively.
The differences between the choice of Ising solvers (simulated annealing, quantum annealing and simulated quenching) and the strategies of the BBO algorithms (BOCS, FMQA and their variations) are discussed for further development of the BBO techniques.
\end{abstract}
\begin{document}

\flushbottom
\maketitle
%
%
\thispagestyle{empty}


\section*{Introduction}

In the last decade, there have grown a number of applications of neural networks and machine learning.
Based on the success of the neural network and machine learning applications in the real-world setting, there has also been an increase in the complexity level of the mathematical models.
These models are represented by parameters in large matrices, such as weight matrices in the neural network.
Suppressing the memory size of the model is a rising issue for implementing those applications in the real-world, especially in edge computing.

One of the authors proposed a lossy decomposition scheme of such matrices into a product of an integer matrix and a real matrix (hereinafter referred to as integer decomposition)\cite{Ambai2014}.
The concept of this matrix decomposition is that the integer matrix has a smaller memory footprint compared with the real matrix in floating-point representation.
The integer variable uses only one or two bits for binary or ternary representation, whereas the floating-point variable uses 32 or 64 bits.
Thus, the compression rate is approximated by the ratio of the sizes of the original matrix and the real matrix.
For a specific machine learning task, the memory footprint is reduced to 1/3, and the performance is 36.9 times faster than the original one with 1.5\% increase in loss of accuracy.

This process of matrix compression optimises the integer and the real matrices simultaneously.
The problem is a mixed-integer non-linear programming (MINLP) problem, an NP-hard problem.
In addition, the original proposal includes a strategy in a greedy manner to reduce optimisation complexity; however, the greedy method does not reach the best solution in general.

In this paper, we improve the procedure of the integer decomposition through a recently developed black-box optimisation (BBO) technique for  binary variables\cite{Baptista2018,Lepretre2019,Kitai2020}.
These methods use a data set of binary input and real-value output to produce a surrogate model, i.e. a pseudo-Boolean function, and optimise the model using Ising solvers.
The tentative optimal solution from the Ising solver can be evaluated by the black-box function, e.g. conducting experiments, simulations or other methods depending on the problem.
In the matrix decomposition, we transform our MINLP problem into a non-linear integer programming (NLIP) problem.
Although we still have an explicit form of the cost function with integer variables, we can consider this function as a black-box function.
Although this paper demonstrates a binary variable problem, it is straightforward to handle an integer variable with binary variables.

Data acquisition is not a hard task thanks to the explicit form of the function.
Thus, the number of iterations and evaluations of the cost function is not a big deal in this case.
We will conduct $O(n^2)$ iterations for BBO, which is a unique feature of this study from a typical case with an expensive-to-evaluate function.

Several tasks on lossy data compression have been commonly used in image and audio data processing, such as JPEG and MP3.
Their algorithms are based on human limitations of frequency and time domains.
General-purpose algorithms, such as low-rank approximation, non-negative matrix factorisation and non-negative/binary matrix factorisation algorithms, have been proposed\cite{Eckart1936,Lee1999,Malley2018}.
These methods are not applicable for the integer decomposition because they target the decomposition by real or non-negative matrix.

Recently, a general matrix decomposition method has been proposed\cite{Yoon2021}.
The proposed method optimises integer matrix and real matrix separately, while our algorithm does simultaneously.
In addition, Thompson sampling of the MINLP surrogate model is studied with commercial mixed-integer programming (MIP) solvers\cite{Daxberger2020}.
Our algorithm provides solutions to MIP problems using Ising solvers, whereas such commercial solvers do not disclose details of algorithms.

A number of studies on the applications of the BBO of binary variables with Ising solvers are carried out, including aero-structural problem\cite{Baptista2018}, nanomaterials\cite{Kitai2020}, spin glass\cite{Koshikawa2021}, commonality optimisation\cite{Koshikawa2021b} and chemical structure optimisation\cite{Hatakeyama2021,Gao2021}.
This paper not only introduces another application but also proposes a novel algorithm for solving MINLP problems by the BBO using the Ising solver.

This paper is organised as follows.
The ``Integer decomposition'' section defines the integer decomposition and derives a black-box formulation of the problem.
Variations of BBO algorithms and Ising solvers tested in this study are described in the ``Black-box optimisation'' section.
The ``Results'' section dedicates to the results of these algorithms and solvers.
The final section summarises and discusses the results.

\section*{Integer decomposition}
\label{sec2}
\subsection*{Original formulation of the integer decomposition}
\label{id_original}

The integer decomposition approximates a target matrix $\bm{W}$ by a product of binary (or ternary) matrix $\bm{M}$ and real matrix $\bm{C}$,
\begin{equation}
\bm{W} \sim \bm{V} = \bm{MC}.
\end{equation}
The matrix sizes of $\bm{W}$, $\bm{M}$ and $\bm{C}$ are $N \times D$, $N \times K$ and $K \times D$, respectively.
This decomposition is parameterised by $K$, which controls the approximation accuracy.

If $K = N$ (no compression), $\bm{M}$ and $\bm{C}$ can be $\bm{I}$ and $\bm{W}$,
\begin{equation}
\bm{V} = \bm{M}\bm{C} = \bm{I}\bm{W} = \bm{W}.
\end{equation}
In the matrix compression, $K$ is smaller than $N$.
Thus, we cannot reproduce $\bm{W}$ by $\bm{M}$ and $\bm{C}$.
Then, the optimal binary and real matrices can be obtained through minimisation of the cost function expressed by $L_2$ matrix norm,
\begin{equation}
\label{eq_argmin_cost_func}
\argmin_{\substack{\bm{M} \in \{-1,1\}^{N \times K}\\\bm{C} \in \mathbb{R}^{K \times D}}} \left\|\bm{W}-\bm{MC}\right\|_2^2.
\end{equation}
This is an MINLP problem; therefore, there is no general algorithm to efficiently find the exact solution.

The original algorithm finds the decomposition as a series of products,
\begin{equation}
\label{eq_Vmc}
\bm{V} = \sum_i^K {\bm{m}_i\bm{c}_i^\mathsf{T}},
\end{equation}
where $\bm m_i$ and $\bm c_i$ are $N$ and $D$ dimensional vectors and calculated step by step from $i=1$ to $i=K$.
The $i$-th vectors $\bm{m}_i$ and $\bm{c}_i$ are optimised by using pre-optimised vectors $\bm{m}_j$ and $\bm{c}_j$ $(j=1,\cdots,i-1)$,
\begin{equation}
\argmin_{\substack{\bm{m}_i \in \{-1,1\}^N\\\bm{c}_i \in \mathbb{R}^D}} \left\|\left(\bm{W}-\sum_j^{i-1}{\bm{m}_j\bm{c}_j^\mathsf{T}}\right)-\bm{m}_i\bm{c}_i^\mathsf{T}\right\|_2^2.
\end{equation}
The search space is drastically reduced from $NK$- to $N$-dimension, although convergence to the exact solution is not guaranteed.
In each step, this algorithm finds the best rank-one approximation of the residual from the previous approximation.
Therefore, updating the variables that are fixed previously is not possible, i.e. it cannot escape from local minima.

As shown in Eq.~\eqref{eq_Vmc}, the matrix decomposition has two types of arbitrariness.
The first one is the order of columns (or rows) in the binary matrix $\bm{M}$ (the real matrix $\bm{C}$) indexed by $i$, that is, the order of sum in Eq.~\eqref{eq_Vmc}.
The second one is the sign of each column in $\bm{M}$ (i.e. $\bm{m}_i, \bm{c}_i \rightarrow -\bm{m}_i, -\bm{c}_i$).
Consequently, the total number of equivalent matrices is $K! \times 2^K$.
In the case of $K=3$, there are $48$ exact solutions of Eq.~\eqref{eq_argmin_cost_func}.
Figure~\ref{fig_s1} in the appendix shows examples of the 48 solutions for a specific instance.

\subsection*{Black-box formulation of the integer decomposition}
\label{id_bbo_formulation}

As shown in the previous subsection, the integer decomposition is an MINLP problem.
To solve this problem by BBO algorithms, we convert it to an NLIP problem.
If given $\bm{M}$ has linearly independent columns, we can calculate $\bm{C}$ using the least-squares method in matrix form,
\begin{equation}
\bm{C} = \left(\bm{M}^\mathsf{T}\bm{M}\right)^{-1}\bm{M}^\mathsf{T}\bm{W},
\end{equation}
where $\left(\bm{M}^\mathsf{T}\bm{M}\right)^{-1}\bm{M}^\mathsf{T}$ refers pseudoinverse, and thus the approximated matrix $\bm{V}$ is a function of $\bm{M}$
\begin{equation}
\bm{V}(\bm{M}, \bm{C}) = \bm{M}\bm{C} = \bm{M}\left(\bm{M}^\mathsf{T}\bm{M}\right)^{-1}\bm{M}^\mathsf{T}\bm{W} = \bm{V}(\bm{M}).
\end{equation}
Substituting this equation to Eq.~\eqref{eq_argmin_cost_func}, we have
\begin{equation}
\label{eq_argmin_m}
\argmin_{\bm{M} \in \{-1,1\}^{NK}} \left\| f(\bm{M}) \right\|_2^2,
\end{equation}
where
\begin{equation}
f(\bm{M}) = \bm{W} - \bm{M}\bm{C} = \bm{W} - \bm{M}\left(\bm{M}^\mathsf{T}\bm{M}\right)^{-1}\bm{M}^\mathsf{T}\bm{W}.
\end{equation}

Now, we remove real-value parameters and have an NLIP problem.
Note that the Taylor series of this cost function has infinite terms, and thus Ising solvers, including quantum annealer, cannot solve this optimisation problem directly.
On the other hand, BBO can solve this problem if we deal with the input-output relationship of the cost function $\left\| f(\bm{M}) \right\|_2^2$ as a black-box function.
The optimisation algorithm employs not the explicit form of the function but a specific data set of the input-output relationship calculated from the function.

\section*{Black-box optimisation}
\label{sec3}
\subsection*{BBO algorithms}

Ising solvers find a solution of a quadratic function so we can approximate the data set $(\bm{x}, y)$ by the quadratic function $\hat{y}(\bm{x}) = \bm{x}^\mathsf{T}\bm{A}\bm{x} + \bm{b}^\mathsf{T}\bm{x} + c$, where $\bm{x} \in \{-1,1\}^{NK}$ and $\bm{A}$, $\bm{b}$ and $c$ are model parameters.
Note that we use $\bm{x}$ in this surrogate model instead of $\bm{M}$, $y$ as the cost associated with a given $\bm{x}$ and $\hat{y}$ as the approximated cost by the surrogate model.
This function can be expressed in a simplified quadratic form $\bm{x}^{\mathsf{T}}\bm{A}\bm{x}$ if we include an additional dimension in the vector $\bm{x}$ as $(x_1, \cdots, x_{NK}, 1)$.
The different algorithm handles the surrogate model differently.

Bayesian optimisation of combinatorial structures (BOCS)\cite{Baptista2018} treats the parameter $\bm{A}$ in Bayesian linear regression.
The authors proposed to use the horseshoe prior\cite{Carvalho2010},
\begin{eqnarray}
\alpha_k | \beta^2_k, \tau^2, \sigma^2 & \sim & N(0,\beta^2_k\tau^2\sigma^2) \nonumber\\
\tau, \beta_k & \sim & C^+(0,1) \\
P(\sigma^2) & = & \sigma^{-2},\nonumber
\end{eqnarray}
where $\alpha_k$ stands for the coefficient of the $k$-th variable in the linear regression and $C^+(0,1)$ is a half-Cauchy distribution.
Note that as the surrogate model is linear, second-order terms $x_i x_j$ are treated as independent explanatory variables, i.e. $(x_1, \cdots, x_n, x_1x_2, x_1x_3, \cdots, x_{n-1}x_n)$, where $n = NK$.
Thus, the index $k$ runs from $1$ to $n+n(n-1)/2$.
As the parameter $\bm{A}$ of the surrogate model is a distribution in BOCS, a specific value of $\bm{A}$ is chosen based on the distribution inspired by Thompson sampling\cite{Thompson1933}.
In addition to the horseshoe prior, normal prior $\alpha_k \sim N(0, \sigma^2)$ and normal-gamma prior $\alpha_k, \sigma^{-2} \sim \text{NormalGamma}(0, 1, 1, \beta)$ are tested.

The sampling from the horseshoe distribution is performed using the Monte Carlo sampling\cite{Carvalho2010}, which requires a longer execution time compared to the normal and normal-gamma distributions.
Samplings of the variables from these distribution functions are accelerated using fast Gaussian samplers\cite{Rue2001,Bhattachrya2016}.

Factorisation machine with quantum annealing (FMQA)\cite{Kitai2020} utilises the factorisation machine (FM)\cite{Rendle2010} as the surrogate model.
The surrogate model of degree $d = 2$ is defined as
\begin{equation}
\hat{y}(\bm{x}) := w_0 + \sum_{i=1}^nw_i x_i + \sum_{i=1}^n\sum_{i=i+1}^n\langle\bm{v}_i,\bm{v}_j\rangle x_i x_j,
\end{equation}
where $\langle\cdot,\cdot\rangle$ represents the dot product of two vectors of size $k_\text{FM}$,
\begin{equation}
\langle\bm{v}_i,\bm{v}_j\rangle := \sum_{l=1}^{k_\text{FM}} v_{i,l} v_{j,l}.
\end{equation}

The horseshoe prior and FM introduce sparsity in the surrogate model, whereas normal and normal-gamma priors do not.
The results section shows the effects of this sparsity on the performances of algorithms.
Each algorithm has its hyperparameter(s) to be fixed before conducting the BBO, i.e. the variance $\sigma^2$ in the normal prior, the shape parameter $\alpha (=1)$ and the inverse scale parameter $\beta$ in the normal-gamma prior and the size parameter $k_\text{FM}$ in the FM, while the horseshoe prior has no hyperparameters.
Hyperparameters $\sigma^2$ and $\beta$ are optimised for a specific instance.
Then, the optimal values are applied to other instances.
We do not optimise the size parameter $k_\text{FM}$ but choose eight from FMQA's proposal as well as 12 to have enough degree of freedom to represent $\alpha_k$.

We refer to the vanilla BOCS as vBOCS, the normal prior BOCS as nBOCS and the normal-gamma BOCS as gBOCS.
The differences in $k_\text{FM}$ are identified by FMQA08 and FMQA12.
In addition, a random search algorithm is referred to as RS, in which each vector $\bm{x}$ is randomly sampled without utilising the pre-obtained data set.

\subsection*{Ising solvers}

BOCS and FMQA use an Ising solver to find the solution of the surrogate model represented in a quadratic form.
We evaluate three Ising solvers: simulated annealing, QA and simulated quenching.
Simulated annealing (SA) introduces a thermal fluctuation in the exploring process of the cost function\cite{Kirkpatrick1983}.
It is implemented on the Monte Carlo simulation.
The temperature in the simulation is a parameter that controls the probability of variable reversal such that the cost function increases.
Initially, the temperature is high to help global search in the solution space and find the candidate basin harbouring the global minimum.
Later, the temperature becomes low to find the lowest solution in the basin. 
If the cooling schedule is slow enough, i.e. $\propto 1/\log(t)$, SA finds the global minimum\cite{Geman1984}.

Quantum annealing (QA) takes place in the thermal fluctuation in SA by quantum fluctuation through a scheduled transverse field.
The quantum system starts from a trivial state (superposition of all solutions) and finds the ground state of the cost function at the end of the annealing\cite{Kadowaki1998}.
If the transverse field is scheduled as $\propto t^{-1/(2N-1)}$, the system converges to the ground state at the end\cite{Morita2007}.

We refer to simulated quenching (SQ) as a variation of SA with extremely rapid quenching of the temperature from high to zero.
Although this algorithm simplifies and accelerates the Monte Carlo calculation, it eliminates the ability to search solution space globally at the early stage of annealing.
Thus, this algorithm tends to be trapped in local minima more frequently than SA and QA.

We utilise SA and QA solvers in D-Wave Ocean SDK with default parameters.
The default initial and final temperatures for SA are determined from approximately estimated maximum and minimum effective fields with scaling factors 2.9 and 0.4, respectively.
The default annealing time for QA is 20 microseconds.
In SQ, the temperature is not annealed but kept constant at 0.1.
We optimise the surrogate model 10 times in each iteration step for all solvers to find a better solution.

\section*{Results}
\label{sec4}

Integer decomposition was conducted for ($N \times D =$) $8 \times 100$ matrix $\bm{W}$, which is constructed by shrinking from the final fully connected layer of VGG16 convolutional neural network\cite{Simonyan2015}.
We choose the decomposition parameter $K = 3$, and thus the matrix $\bm{W}$ is decomposed into a $8 \times 3$ binary matrix $\bm{M}$ and a $3 \times 100$ real matrix $\bm{C}$.
As discussed in ``Original formulation of the integer decomposition'' subsection, the size $n$ of the problem is determined by the size of the binary matrix.
In this case, $n = 8 \times 3 = 24$.
The dimension of the model parameter $\alpha_k$ is $O(n^2)$ so that we start from the initial data set of size $n$ and then add $2n^2$ data points iteratively to reach enough size for estimating the model parameter. 
In the current analysis ($n = 24$), we generate initial 24 data points followed by 1,152 iterations (1,176 in total).
This number is remarkably small compared to the solution space {$2^{24} \sim$} $1.7\times10^7$.
All the algorithms we test in this paper are randomised algorithms and/or starting from a randomly selected data set, and thus we conduct 25 runs (or 100 runs for RS) to estimate the average performance.
The performance also depends on the matrix $\bm{W}$.
Therefore, we generate ten individual problem matrices (instances) to evaluate the performance over instances.

Figure 1 shows the results from different algorithms for the first instance.
In the BBO process, SA is used as an Ising solver.
The x and y axes represent the iteration step, the size of acquired data (including initial data) in linear scale and the residual error from the exact solution in log scale, respectively.
Due to the limitation of the problem size $n$ we tested, the compression quality is insufficient for typical applications.
In this situation, the absolute error is not an appropriate measure to compare algorithms.
Thus, we employ the residual error $\left(\left\| f(\bm{M}) \right\|_2 - \left\| f(\bm{M}^*) \right\|_2\right) / \left\| W \right\|_2$, where $\bm{M}^*$ is the exact solution of Eq.~\eqref{eq_argmin_m}.
With this measure, we can evaluate how the solutions are close to the exact solution. 
The exact and the second-best solutions were separately obtained from brute-force search.
The absolute error of the exact solution $\left\| f(\bm{M}^*) \right\|_2 / \left\| W \right\|_2$, subtracted in the plot, is 0.461.
All algorithms outperform the original approximated solution in the red-dotted line.
The FMQA algorithm improves its solution faster than other algorithms at the early stage.
However, the improvement does not continue at the later stage.
vBOCS and nBOCS improve slowly, but the improvement continues during the process.
They go under the line of the second-best solution (in grey-dotted line), suggesting that some individual runs find the exact solution.

\begin{figure}[thb]
    \centering
    \includegraphics[width=120mm]{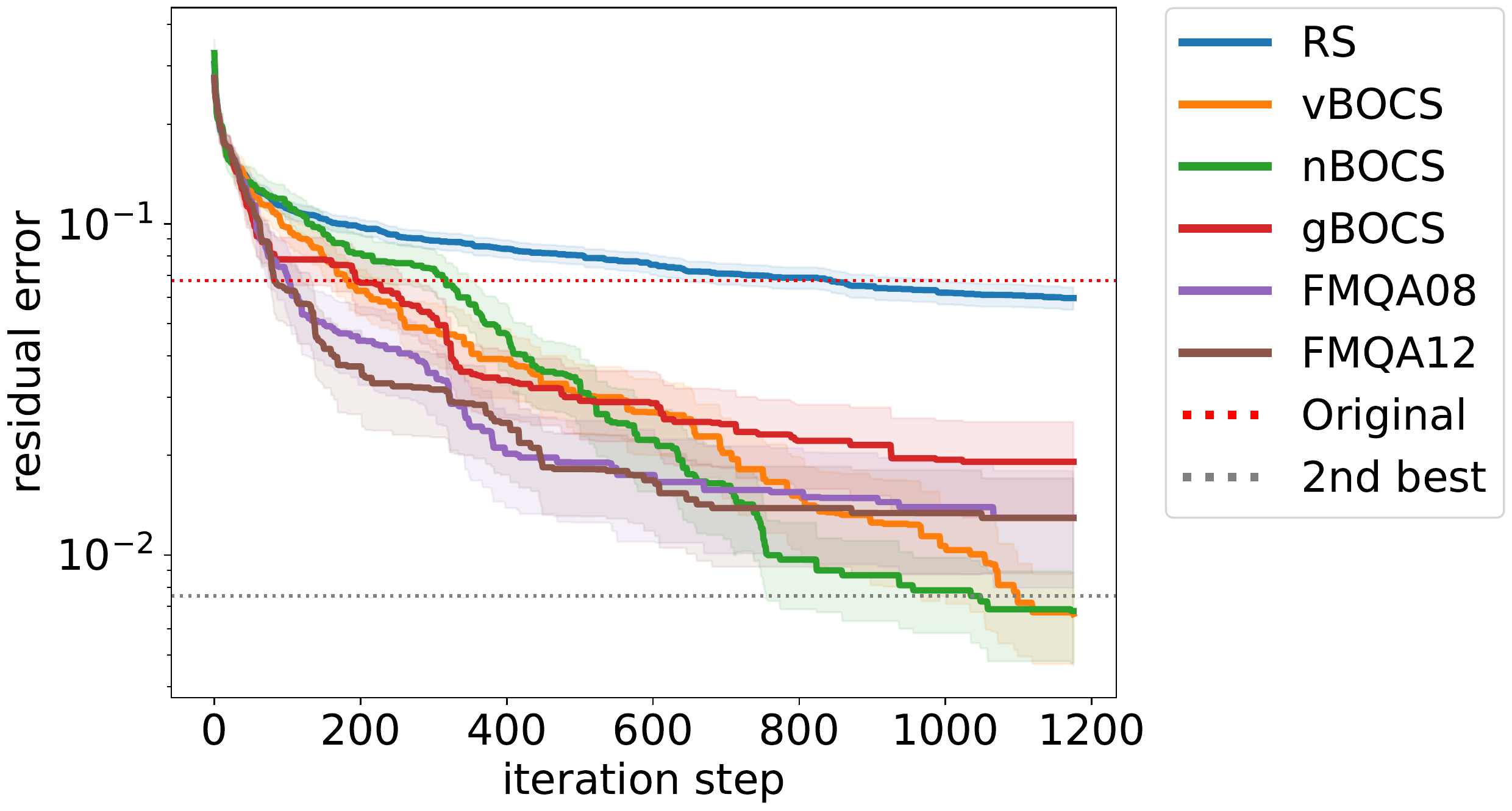}
    \caption{
        \label{fig1}
        The residual error of lossy compression of a matrix (instance) as a function of iteration step among various algorithms with 95\% confidence intervals.        
        The red- and grey-dotted lines indicate the residual error of the original algorithm and the second-best solution by brute-force search.
	}
\end{figure}

Figure 2 shows the differences between Ising solvers (SA, QA and QS) applied to nBOCS, labelled as nBOCS, nBOCSqa and nBOCSsq, respectively.
Although SQ generally has a poor performance in finding a global minimum in a complex landscape of the cost function, there are no clear differences among Ising solvers.
This finding suggests that the landscape of the surrogate model is simple.
Thus, such a simple Ising solver is enough for the current BBO task.

\begin{figure}[thb]
    \centering
    \includegraphics[width=120mm]{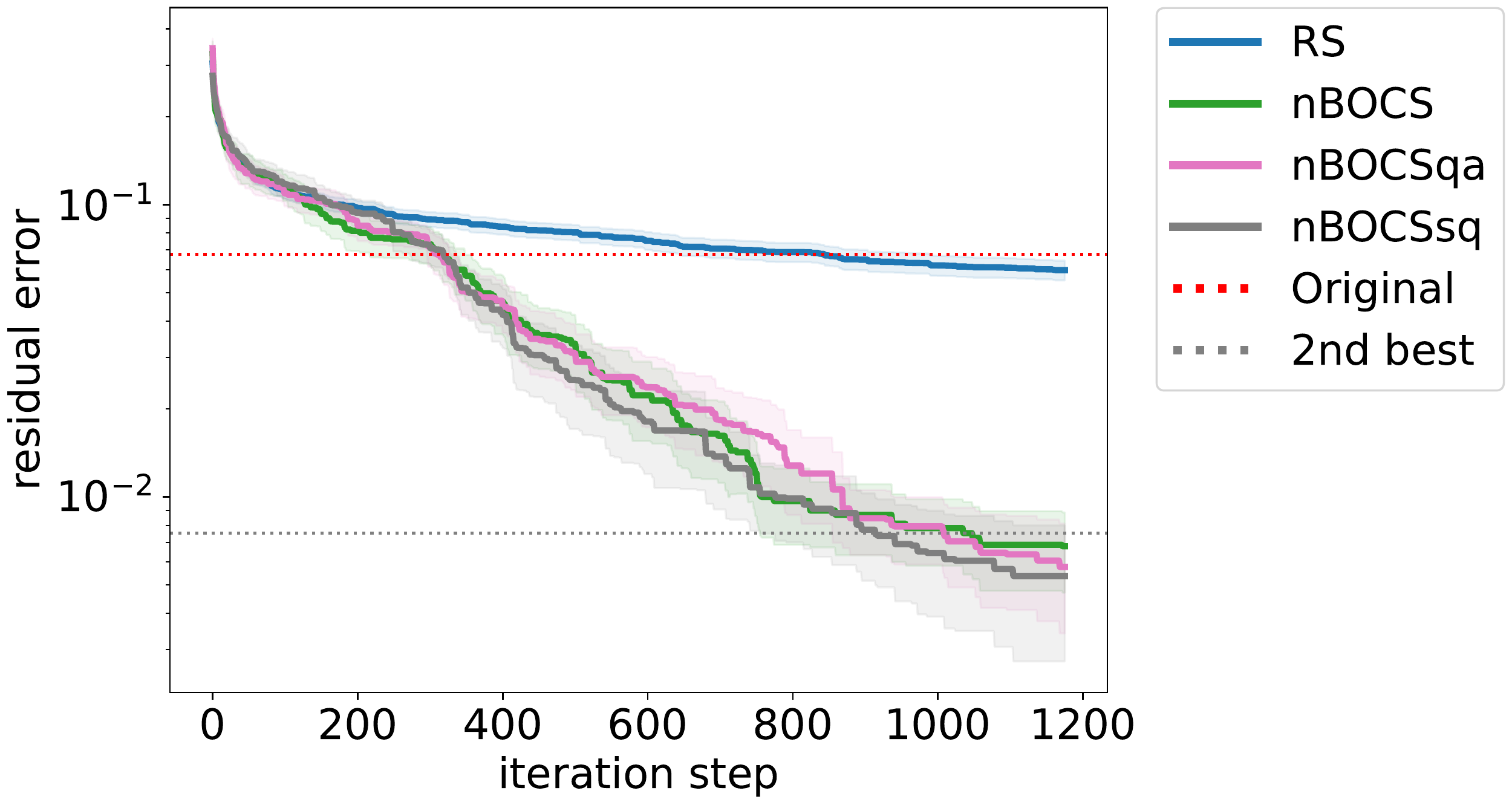}
    \caption{
        \label{fig2}
        The residual error of lossy compression of a matrix as a function of iteration step among the Ising solvers (SA, QA and SQ).
	}
\end{figure}

As discussed in the ``Original formulation of the integer decomposition'' subsection, there are $K! \times 2^K$ exact solutions in the optimisation problem.
In other words, any $\bm{x}$ has $K! \times 2^K - 1$ equivalent vectors, which give the same value of the cost function.
These equivalent vectors can be added to the data set and may accelerate the BBO process.
Figure 3 shows the results of this data augmentation.
The data augmentation (nBOCSa) does not require the additional calculation of the cost function; thus, we do not change the scale of the x-axis.
The results clearly show that the data augmentation negatively affects the performance in the later stage, while there is a little advantage in the early stage.

\begin{figure}[thb]
    \centering
    \includegraphics[width=120mm]{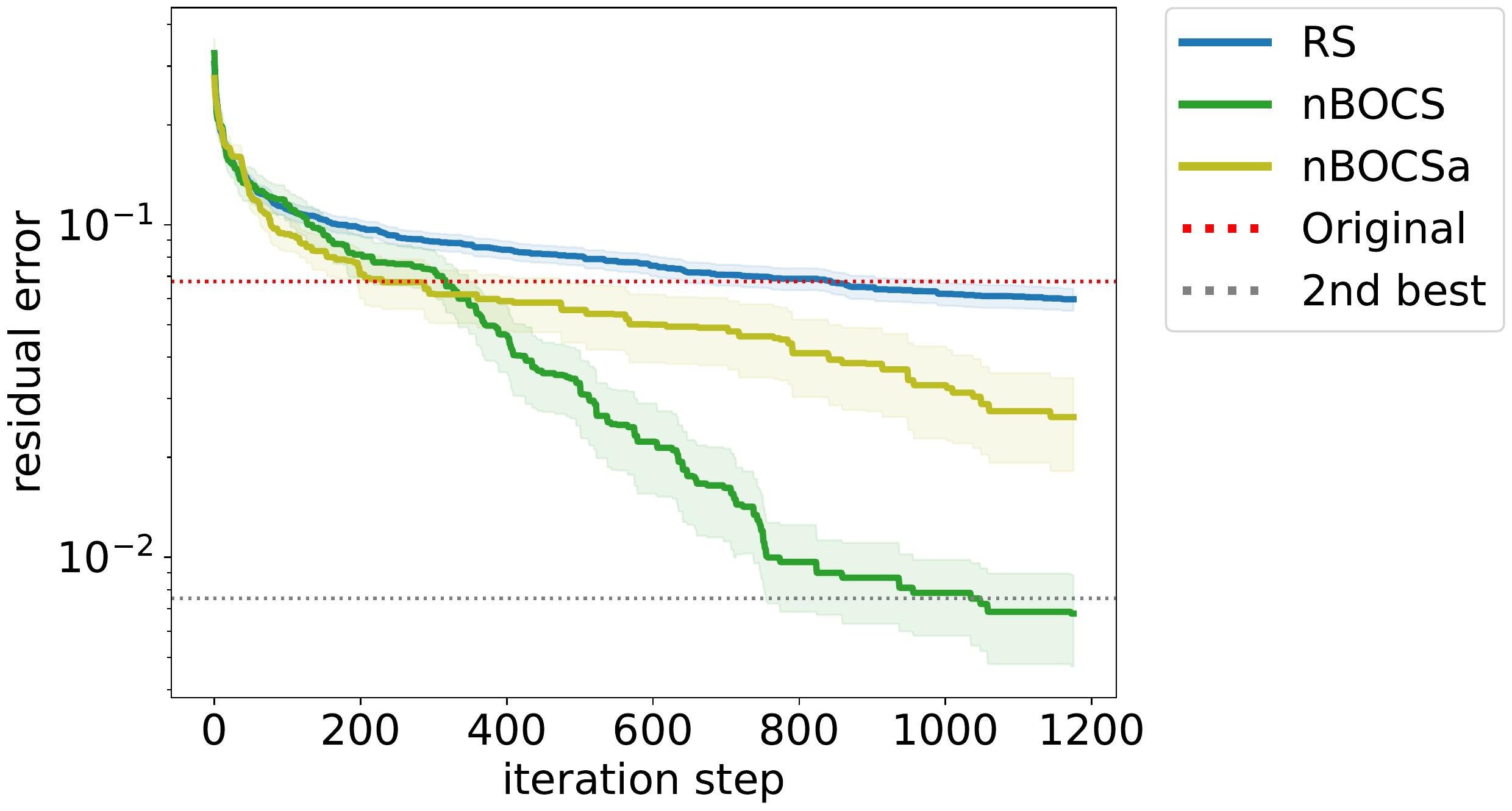}
    \caption{
        \label{fig3}
        The residual error of lossy compression of a matrix for RS and nBOCS variants with and without data augmentation (nBOCSa and nBOCS).
	}
\end{figure}

Table~\ref{table1} shows the total count of finding exact solutions among all ten instances for all algorithms and their variations.
Algorithms of nBOCS with SA, QA and SQ perform better than others.
RS and nBOCSa cannot find the exact solution in most instances.

\begin{table}[htb]
	\caption{Counts of finding exact soulution per 25 runs. The highest values are in bold.}
	\label{table1}
	\centering
	\begin{tabular}{c|cccccc|cc|c}
		Instance No. & RS & vBOCS & nBOCS & gBOCS & FMQA08 & FMQA12 & nBOCSqa & nBOCSsq & nBOCSa \\
		\hline
		1 & 0 & 7 & 7 & 2 & 1 & 3 & 9 & \textbf{11} & 0 \\
		2 & 1 & 6 & 11 & 5 & 6 & 9 & 12 & \textbf{15} & 1 \\
		3 & 0 & 0 & \textbf{13} & 5 & 6 & 9 & 10 & 2 & 4 \\
		4 & 1 & 16 & \textbf{20} & 10 & 10 & 13 & 14 & 17 & 0 \\
		5 & 0 & 0 & 1 & 0 & 3 & \textbf{4} & 2 & 1 & 0 \\
		6 & 0 & 14 & 12 & 4 & 6 & 4 & 18 & \textbf{20} & 0 \\
		7 & 0 & 1 & 4 & 6 & 7 & \textbf{8} & 7 & 6 & 0 \\
		8 & 1 & \textbf{4} & 2 & 2 & \textbf{4} & 1 & \textbf{4} & 1 & 0 \\
		9 & 1 & 5 & 1 & 3 & 3 & 4 & \textbf{6} & 4 & 0 \\
		10 & 5 & 21 & 20 & 15 & 20 & 21 & 22 & \textbf{25} & 6 \\
		\hline
		Total & 9 & 74 & 91 & 52 & 66 & 76 & \textbf{104} & 102 & 11
	\end{tabular}
\end{table}

\begin{figure}[thb]
    \centering
    \includegraphics[width=160mm]{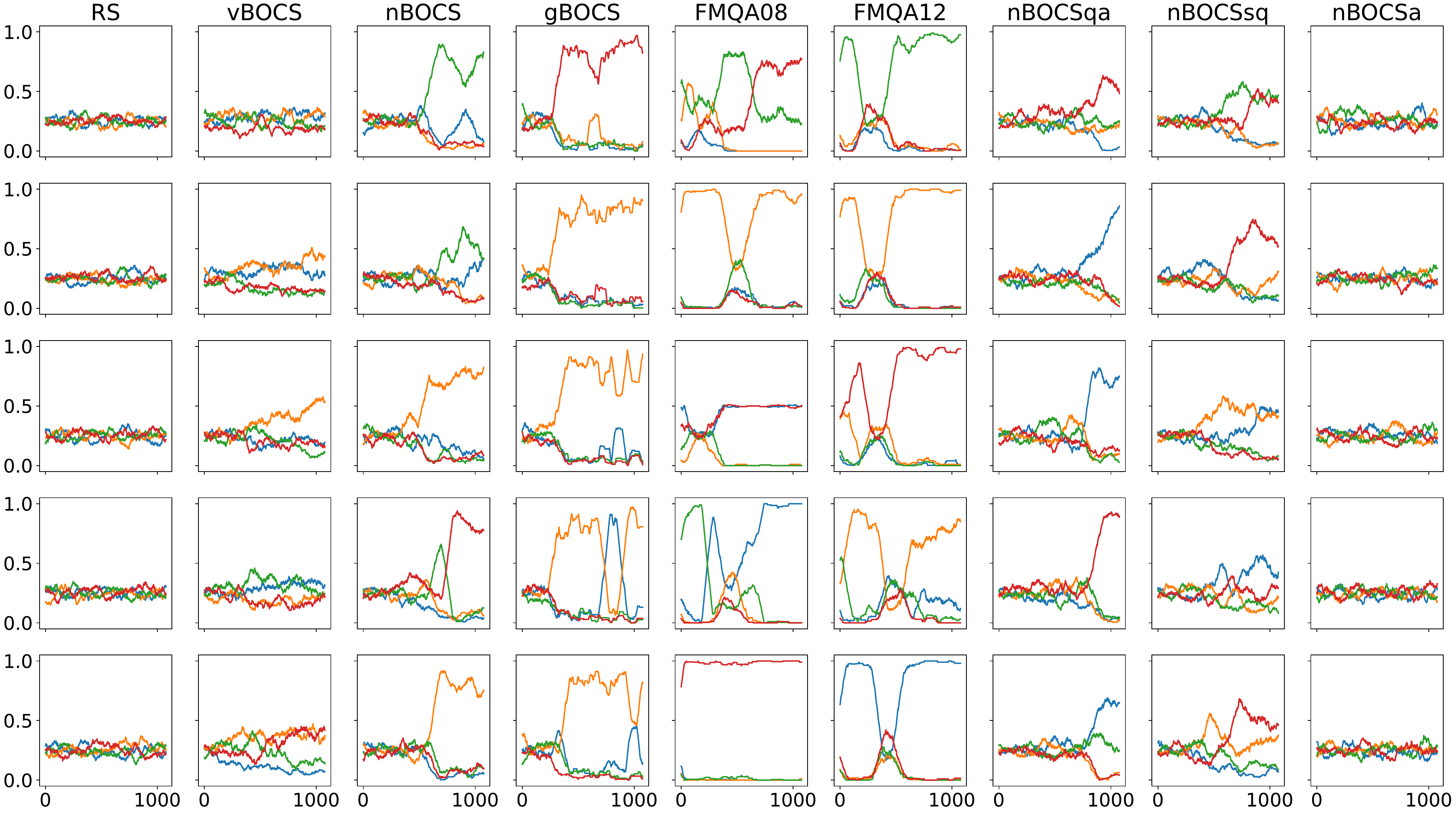}
    \caption{
        \label{fig4}
        The population of the four domains for five individual runs for various algorithms.
        The x-axis represents the iteration step.
	}
\end{figure}

To investigate the behaviour of the algorithms, we perform a cluster analysis of the candidate solution for each step.
If an algorithm gradually focuses on subspace harbouring a specific exact solution, sampling of candidate solutions will also be biased gradually.
We divide the solution space into four domains based on the hierarchical clustering of the exact solutions.
Other solutions in the solution space are assigned to the closest exact solution measured using the Hamming distance and then grouped into the four domains.
The population of the four domains reflects the sampling bias.
Figure~\ref{fig4} shows the population of the candidate solutions among four domains for individual five runs.
The plots are convolved for smoothing (window size, 100). 
RS and nBOCSa show no trend, while FMQAs start focusing on a domain from an early stage.
Other BOCS variants select a domain in the middle of the analysis but continue to explore other domains.

Execution times for all algorithms are shown in Table~\ref{table2}.
nBOCSqa takes a longer time than nBOCS due to the overhead to prepare the data matrix to be uploaded to the quantum annealer.
The communication time between CPU and QPU is not taken into account.
nBOCS is 129 and 67 times faster than vBOCS and FMQA08.
The original algorithm and the brute-force search execution times are 0.00096 and 5553.51 s, respectively.

\begin{table}[htb]
	\caption{Average execution time (s) per run.}
	\label{table2}
	\centering
	\begin{tabular}{c|cccccc|cc|c}
		 & RS & vBOCS & nBOCS & gBOCS & FMQA08 & FMQA12 & nBOCSqa & nBOCSsq & nBOCSa \\
		\hline
		CPU & 0.72 & 7165.06 & 55.39 & 112.39 & 3711.31 & 3625.92 & 241.46 & 55.94 & 319.98 \\
		QPU & - & - & - & - & - & - & 11.60 & - & -
	\end{tabular}
\end{table}

\section*{Discussion}
\label{sec5}

We conducted BBO of the integer decomposition, a lossy compression of a matrix.
Here, we demonstrated that this MINLP problem is transformed into an NLIP problem.
The cost function is pseudo-Boolean because the exploratory variables are binary.
We employed BBO algorithms such as BOCS and FMQA to the cost function.
These algorithms optimise a surrogate model constructed from the input-output data of the cost function to obtain the next candidate for data acquisition and update the model.
This transformation can be generalised to solve MIP problems if the cost function is linear in terms of the real variables.

Among the variations of BOCS and FMQA algorithms, BOCS with normal prior showed the best performance, with an execution time that is one or two digits faster than FMQA or BOCS with horseshoe prior.
FMQA improves the solution rapidly at the initial stage, while BOCS slowly but continuously improves and obtains a superior solution.
The differences in finding the solution space for these algorithms were analysed through clustering.
FMQA tends to focus on a subspace earlier than BOCS.
However, once the algorithm is trapped in a local minimum, the algorithm cannot escape from the local minimum because FMQA is deterministic.
As BOCS is a randomised algorithm, it takes steps but explores larger space and finds a better solution.
In the integer decomposition, the cost function is not an expensive-to-evaluate function; thus, we can conduct enough iterations.
The strength of BOCS in the late stage of the iteration is favourable compared to FMQA.
A randomised version of FMQA\cite{Matsumori2022} should be studied in the future.

In the current formulation, the solution space is divided into equivalent $K! \times 2^K$ subspaces due to the nature of the problem, where each subspace has the exact solution.
We confirmed that the data augmentation for the subspaces did not improve the performance.
BOCS and FMQA approximate the cost function in the quadratic function, which means the surrogate model fits well locally, and not globally.
Therefore, these algorithms try to focus their sampling on a subspace that harbours one of the exact solutions to the problem.
However, data augmentation deals with all the subspaces equally.
Thus, it is impossible to sample from biased solution space to improve the model locally around the exact solution.
Bias is introduced randomly from the selection of initial data and the following biased candidate selections.

With regard to the choice of Ising solvers as a back-end of a black-box optimiser, there is no significant difference between SA, QA and SQ.
This finding is non-trivial because SQ has a poor performance in general.
SQ does not explore the solution space globally because the algorithm only accepts the next solution that lowers the cost.
This fact leads to the following hypothesis.
In the early stage of exploration, the low-quality solutions from the Ising solver are enough to construct a more accurate surrogate model.
Then, in the late stage, the surrogate model approximates the cost function around one of the exact solutions as the sampling is biased.
Optimisation of the surrogate model by the Ising solver might be relatively easy compared to the explicit form of the original cost function.
Therefore, the advantage of QA in BBO is an open question.
Further studies are needed to reveal how the approximated landscape of the cost function learned from the acquired data set changes according to the increase of the data.

While the original algorithm cannot find the exact solution, the proposed BBO-based algorithm can find it with a certain probability in the tested problem size ($n=24$).
However, the execution time of the proposed algorithm takes five digits longer than the original.
(Brute-force search needs additional two digits.)
As the surrogate model is quadratic, BBO needs $O(n^2)$ iterations to estimate the model parameters.
If we employ nBOCSsq (normal prior BOCS with SQ), the most expensive calculation in each iteration is matrix inversion $O(n^3)$ for building the surrogate model.
Therefore, the algorithm takes $O(n^5)$ calculation time.
This scaling may worsen if we choose other Ising solvers (SA and QA).
The proposed algorithm has an advantage in solution quality against the original algorithm and calculation time against the brute-force search.
However, with the current scaling, the typical use of matrix compression, such as weight matrix in a machine learning task, is not applicable.
Figure 4 of the reference \cite{Kitai2020} shows that FMQA finds meaningful solutions for $n = 12 \sim 50$.
Although they fixed the iteration at 2000, more iterations seem needed for large problems.
Further investigation is required to accelerate the calculation for handling the typical size of matrices.

In this paper, the algorithm's limitation is that the model is approximated by quadratic form.
The COMBO algorithm considers higher-order terms by diffusion kernel of graph representation\cite{Oh2019}.
Kernel-based algorithms relaxing the binary variables to the continuous ones are also proposed\cite{Buathong2020,Deshwal2021}.
They may work better for problems where the kernel removes essential difficulties in combinatorial optimisation.
Once the gradient can be calculated, construction and optimisation of the surrogate model can be accelerated\cite{Wu2020}.
Comparison with these algorithms will be a future task.

\section*{Methods}
\subsection*{Shrunk VGG matrix}

To test BBO for lossy matrix compression, we prepare matrices by shrinking the VGG matrix, the weight matrix in the convolutional neural network for image recognition.
We choose the matrix of the final fully connected layer ($4,096 \times 1,000$ matrix).
As the matrix is too large to conduct BBO, we reduce the size by keeping the structural information of the matrix as follows:
the weight matrix $\bm{W}_0$ is decomposed by singular value decomposition,
\begin{equation}
	 \bm{W}_0 = \bm{U} \bm{\Sigma} \bm{V}^\mathsf{T},
\end{equation}
where $\bm{U}$, $\bm{\Sigma}$ and $\bm{V}$ are $4,096 \times 4,096$, $4,096 \times 1,000$ and $1,000 \times 1,000$ matrices, respectively.
Then, we choose eight and a hundred rows/columns from $\bm{U}$ and $\bm{V}$, respectively, and eight singular values from $\bm{\Sigma}$ to construct the shrunk $8 \times 100$ matrix.

\subsection*{Exact solutions}

\begin{figure}[thb]
    \centering
    \begin{overpic}[width=80mm]{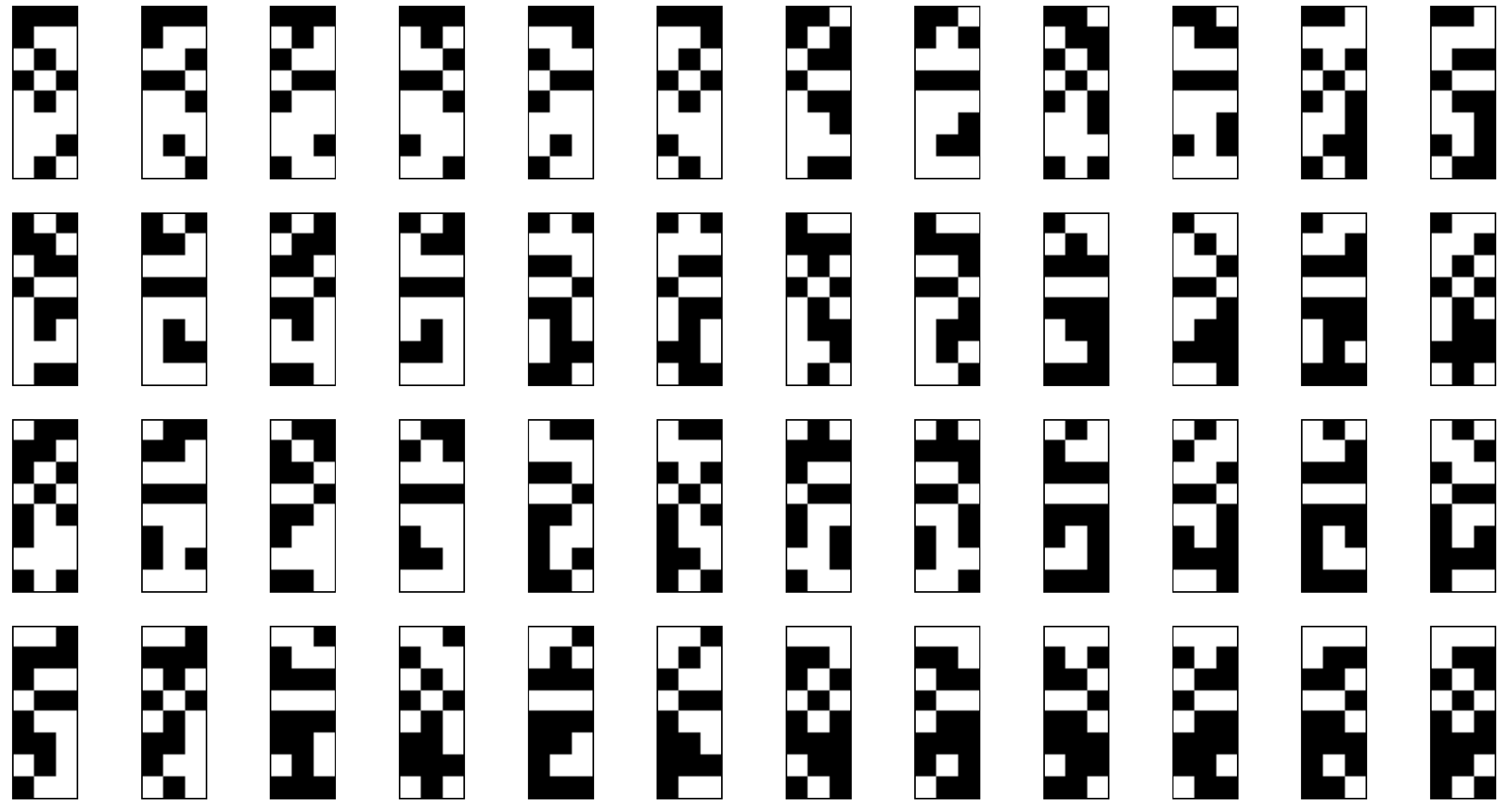}
        \put(-6,50){(a)}
    \end{overpic}
    \begin{overpic}[width=80mm]{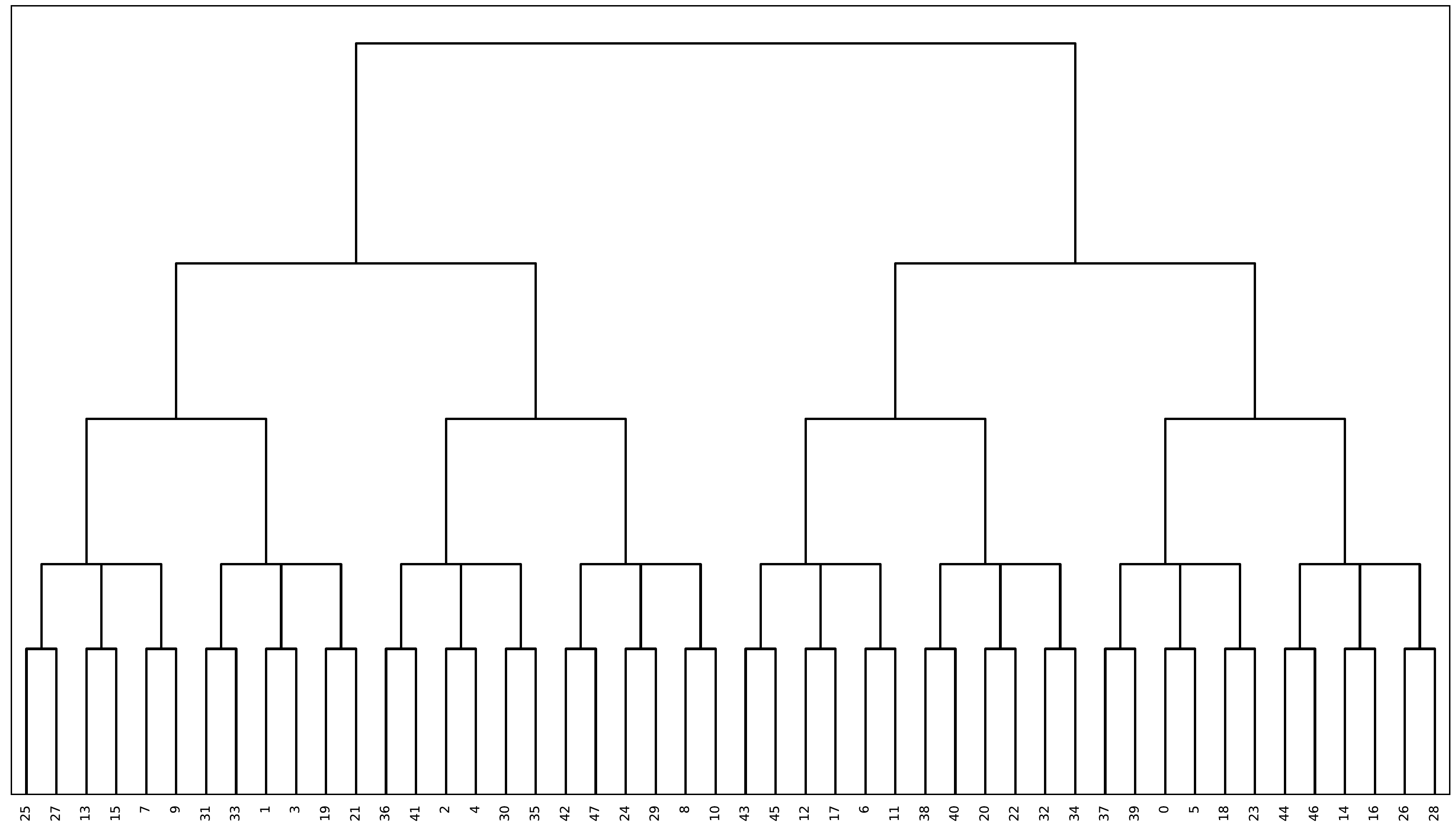}
        \put(3,50){(b)}
    \end{overpic}
    \caption{
        \label{fig_s1}
        (a) 48 exact solutions for the first instance.
        Each box represents a $3 \times 8$ matrix $\bm{M}$.
        Black and white pixels represent $1$ and $-1$, respectively.
        (b) clustering results of the 48 exact solutions by the Ward method.
        Each solution is labelled from 0 to 47 and assigned left-to-right and top-to-bottom in (a).
	}
\end{figure}

The exact solutions in Fig.~\ref{fig_s1}(a) are obtained by solving Eq.~\eqref{eq_argmin_m}.
For the current problem size, we can perform a brute-force search for all candidates in the solution space $2^{24}$.
Each solution is presented in a box of $8 \times 3$ pixels.
Black and white pixels represent $1$ and $-1$, respectively.
Figure~\ref{fig_s1}(b) shows clustering results of the 48 exact solutions by the Ward method.
Similar solutions are grouped, e.g. the second and fourth boxes (labelled 1 and 3 in the cluster) are one of the closest pairs.
We can make four groups by choosing an appropriate cut-off value (the height of the tree).
These four groups are used for colouring in Fig.~\ref{fig4}.

\subsection*{Hyperparameter optimisation}

\begin{figure}[thb]
    \centering
    \includegraphics[width=80mm]{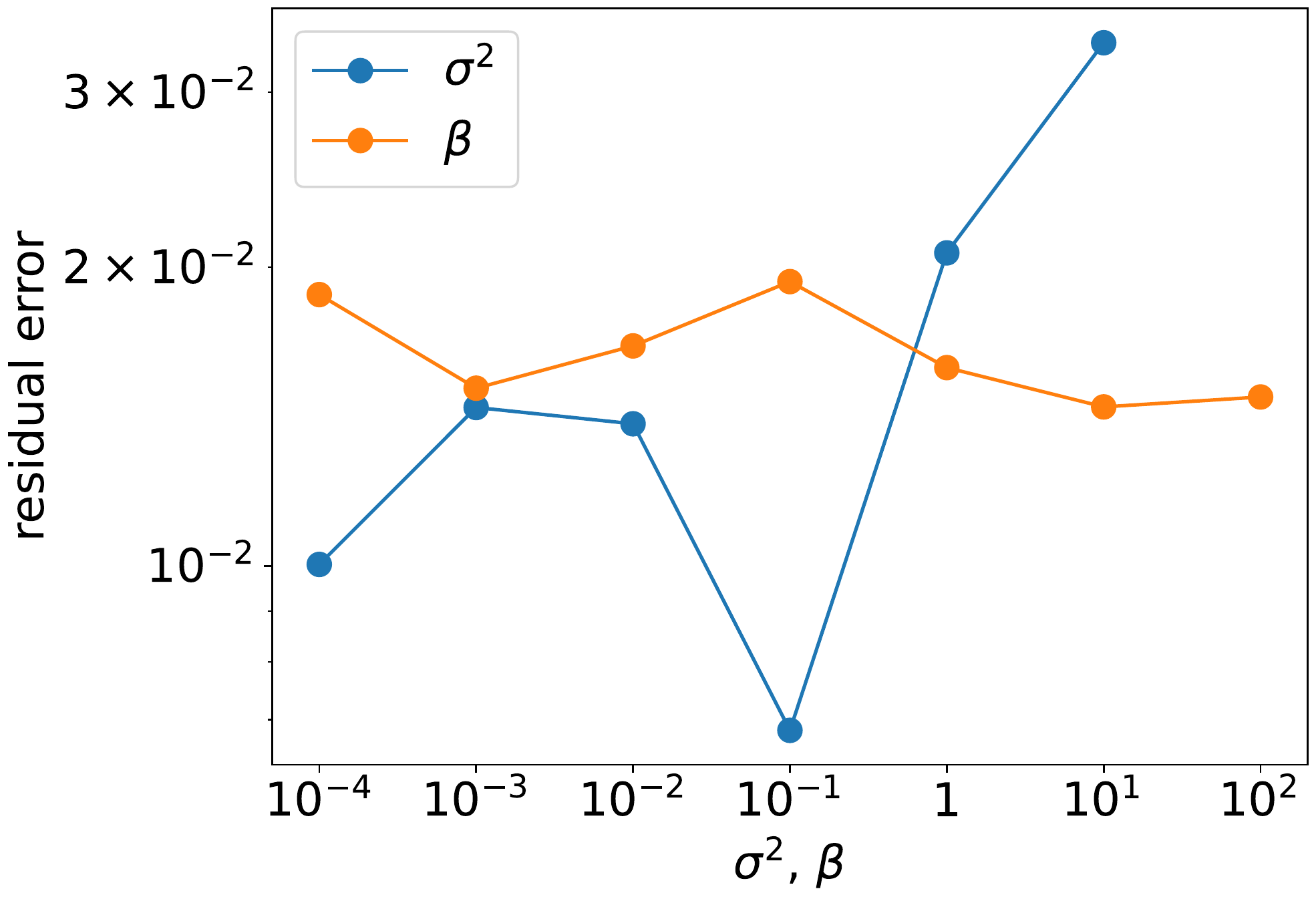}
    \caption{
        \label{fig_s2}
        Hyperparameter dependence on the cost.
	}
\end{figure}

Hyperparameters $\sigma^2$ (variance) for nBOCS and $\beta$ (inverse scale) for gBOCS are optimised for the first instance.
For each hyperparameter, grid search ${0.0001, 0.001, 0.01, 0.1, 1, 10}$ and ${0.0001, 0.001, 0.01, 0.1, 1, 10, 100}$ are conducted (Fig.~\ref{fig_s2}).
In nBOCS, we select $\sigma^2 = 0.1$, which gives the lowest cost.
On the other hand, in gBOCS, we select not $\beta = 10$ but $\beta = 0.001$, as broader prior distribution is preferable for exploring an accurate model, and there is less cost variation among different hyperparameter values.

\subsection*{Results of the other nine instances}

\begin{figure}[thb]
    \centering
    \includegraphics[width=50mm]{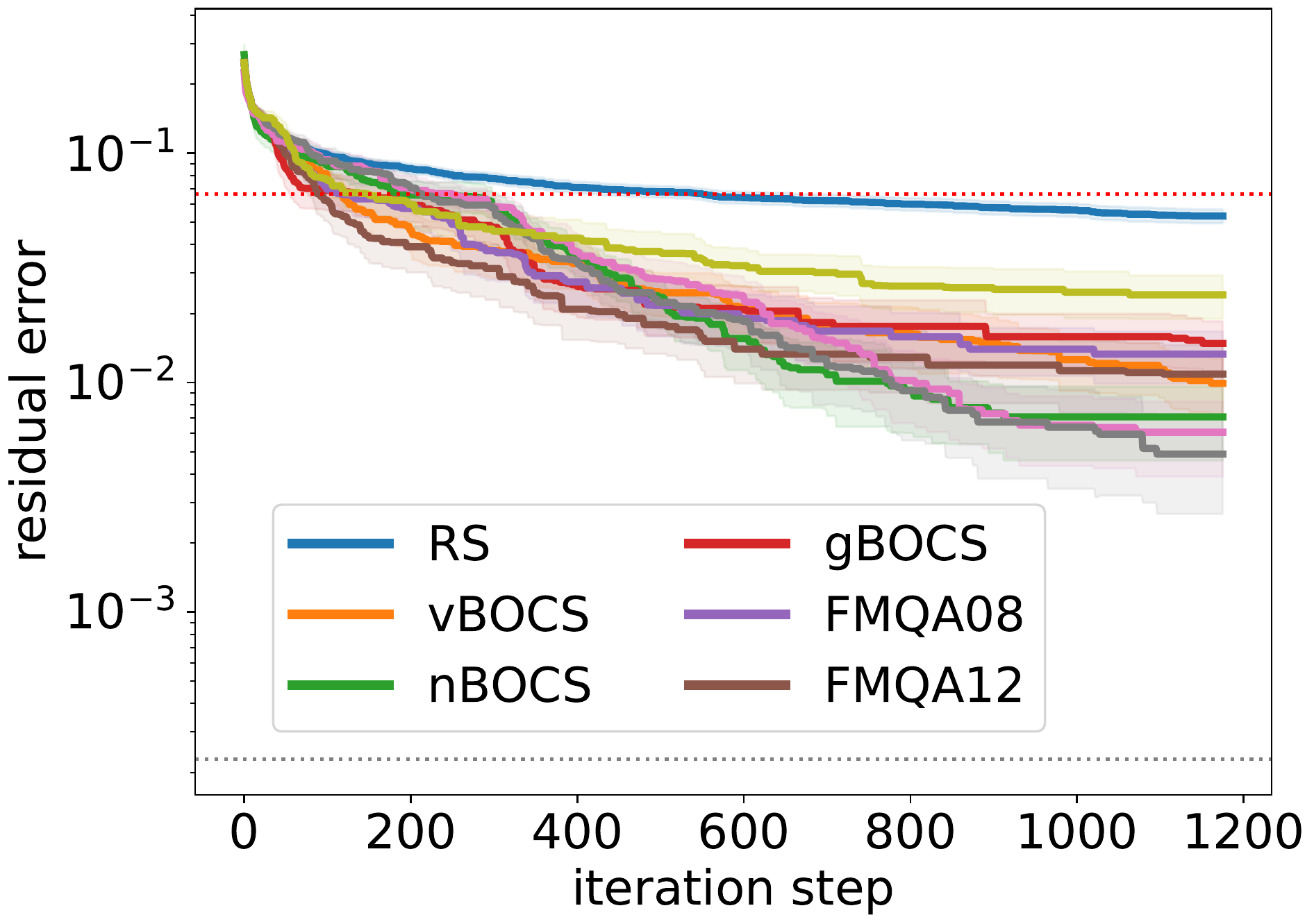}
    \includegraphics[width=50mm]{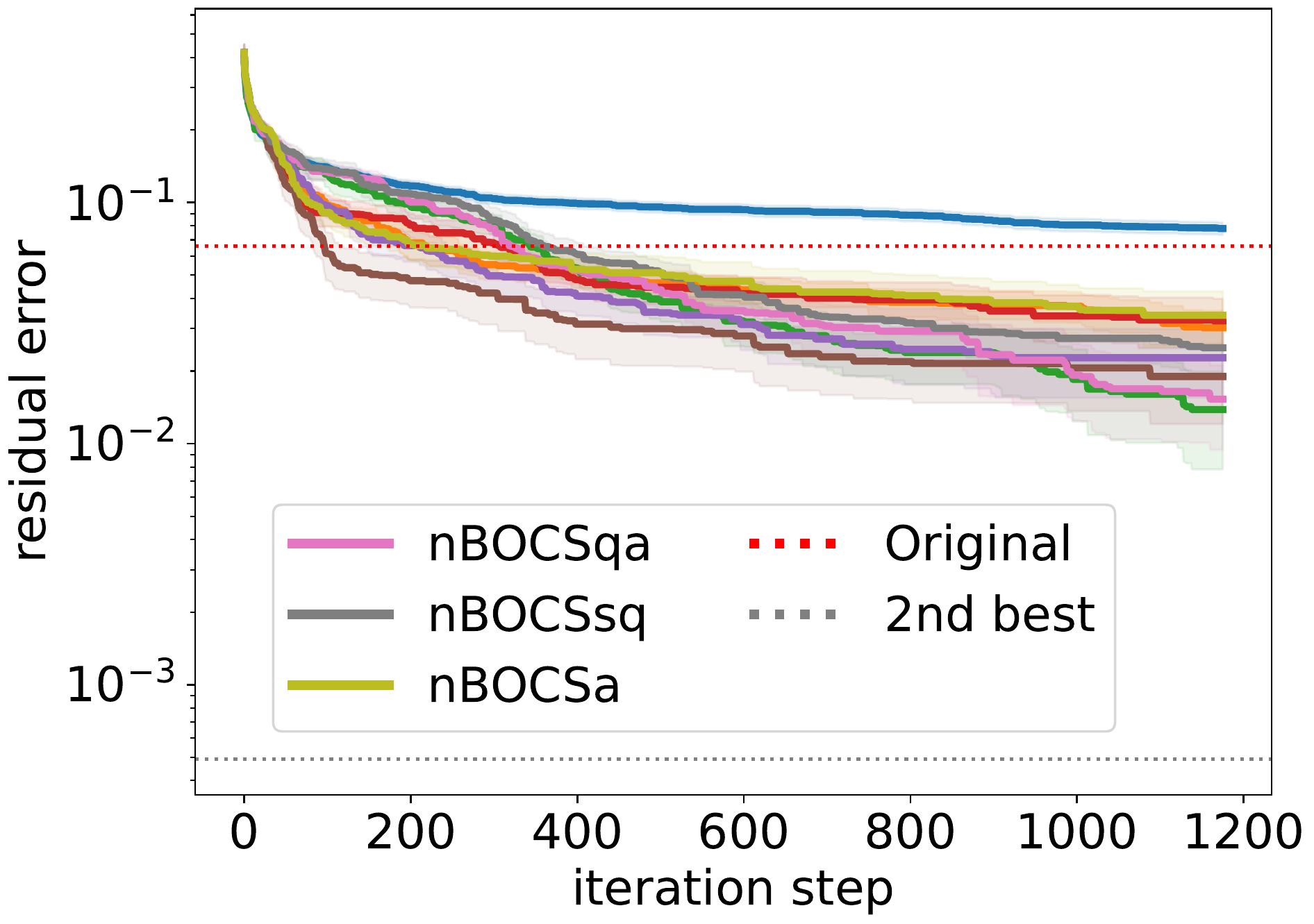}
    \includegraphics[width=50mm]{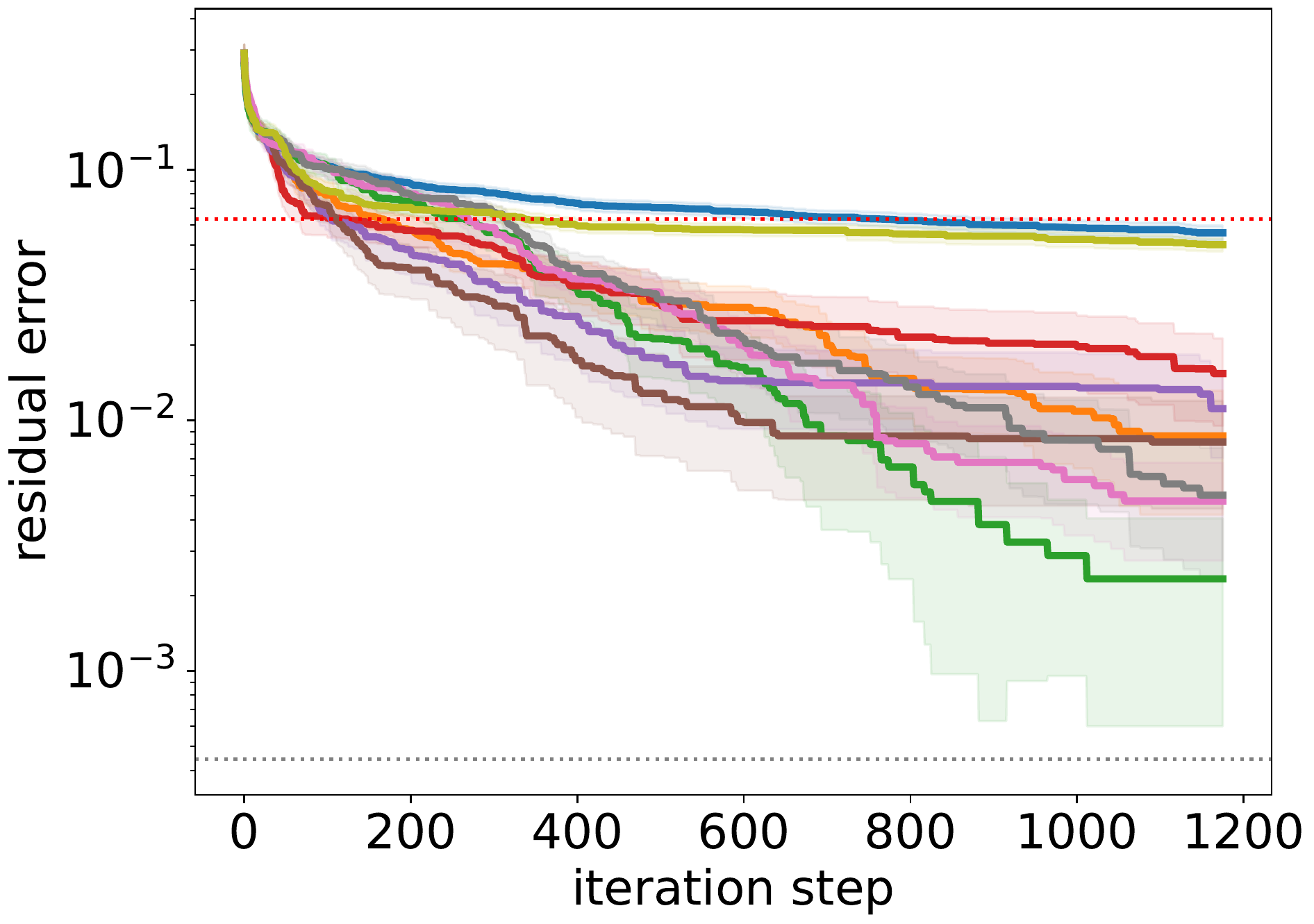}
    \includegraphics[width=50mm]{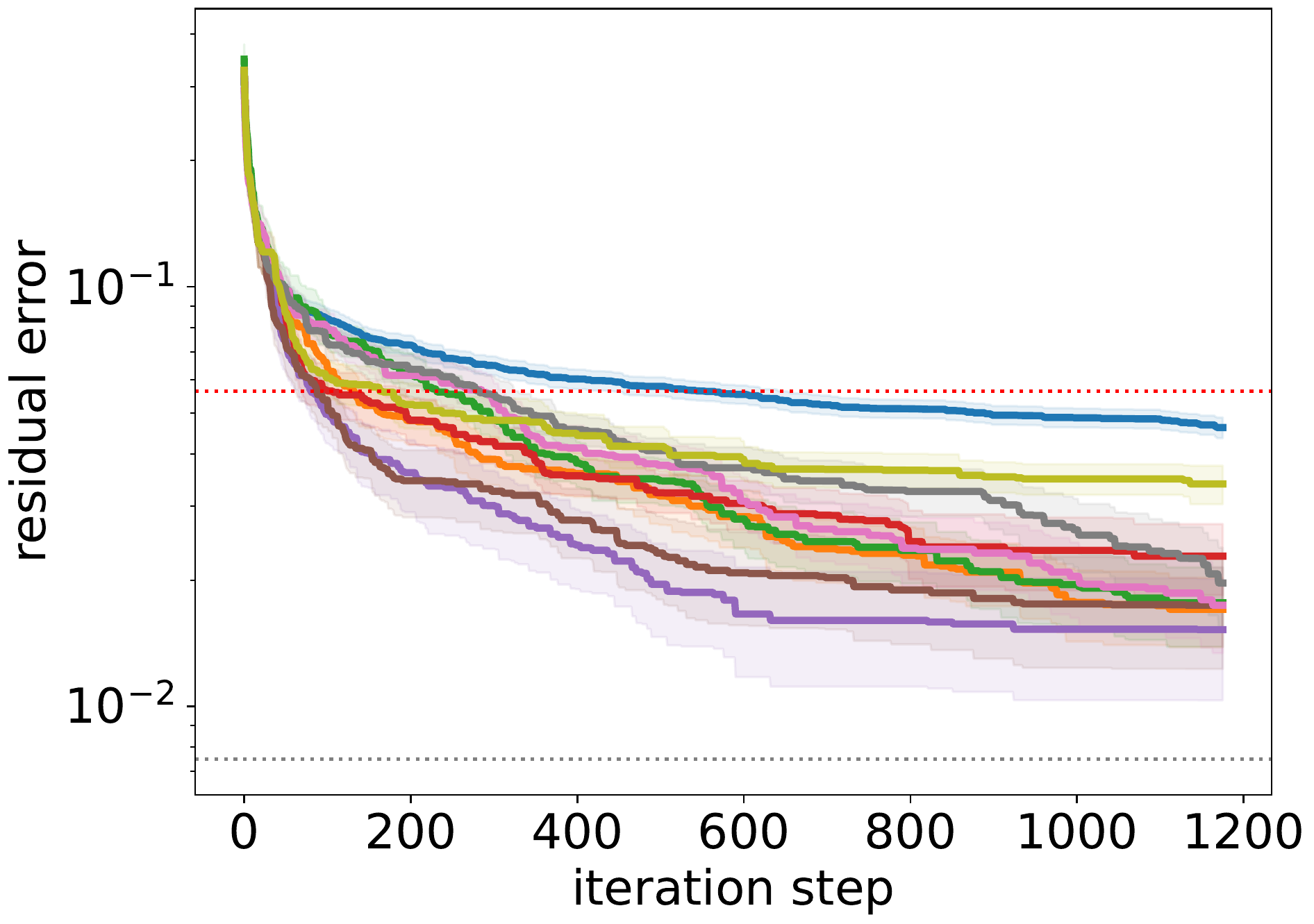}
    \includegraphics[width=50mm]{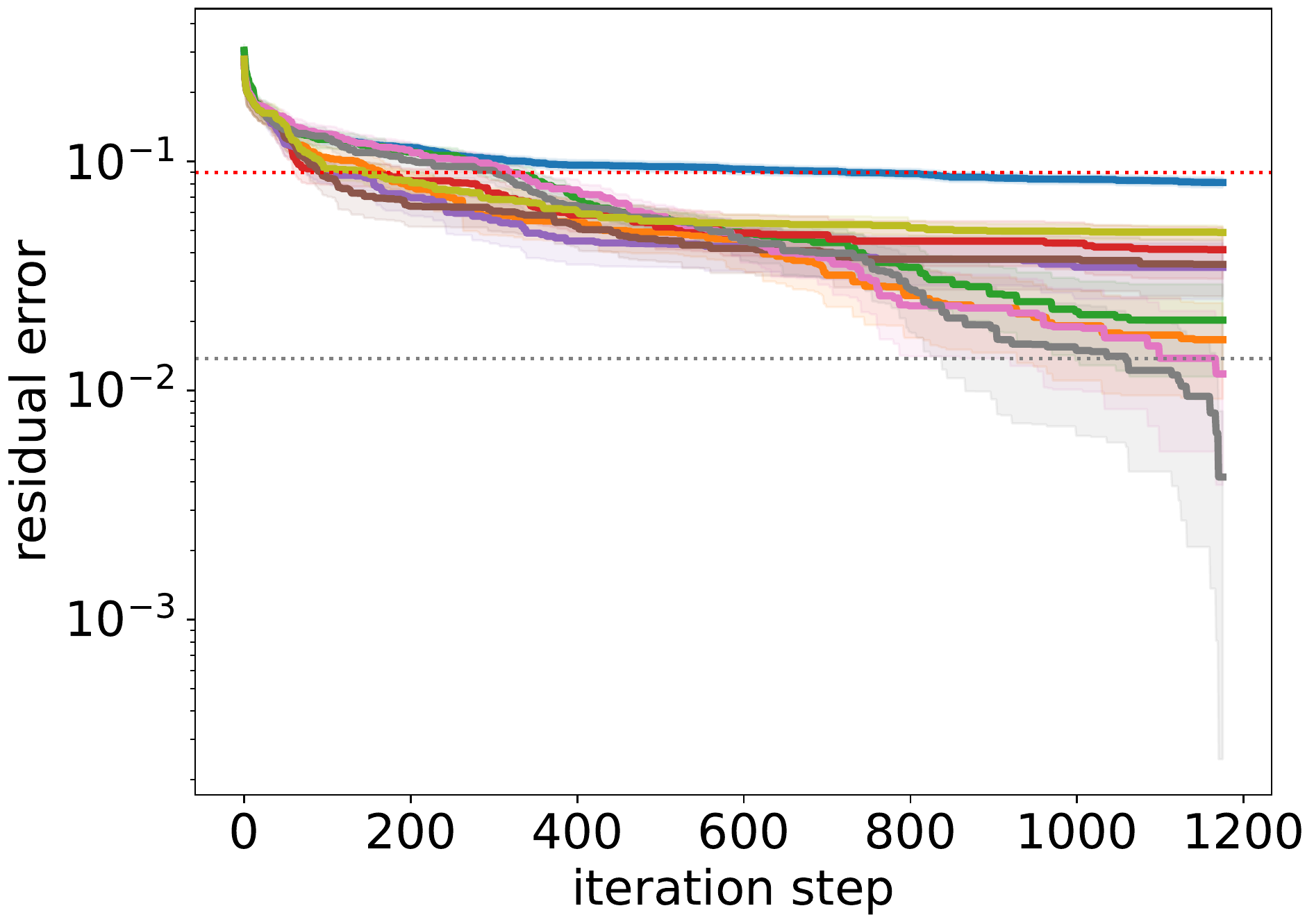}
    \includegraphics[width=50mm]{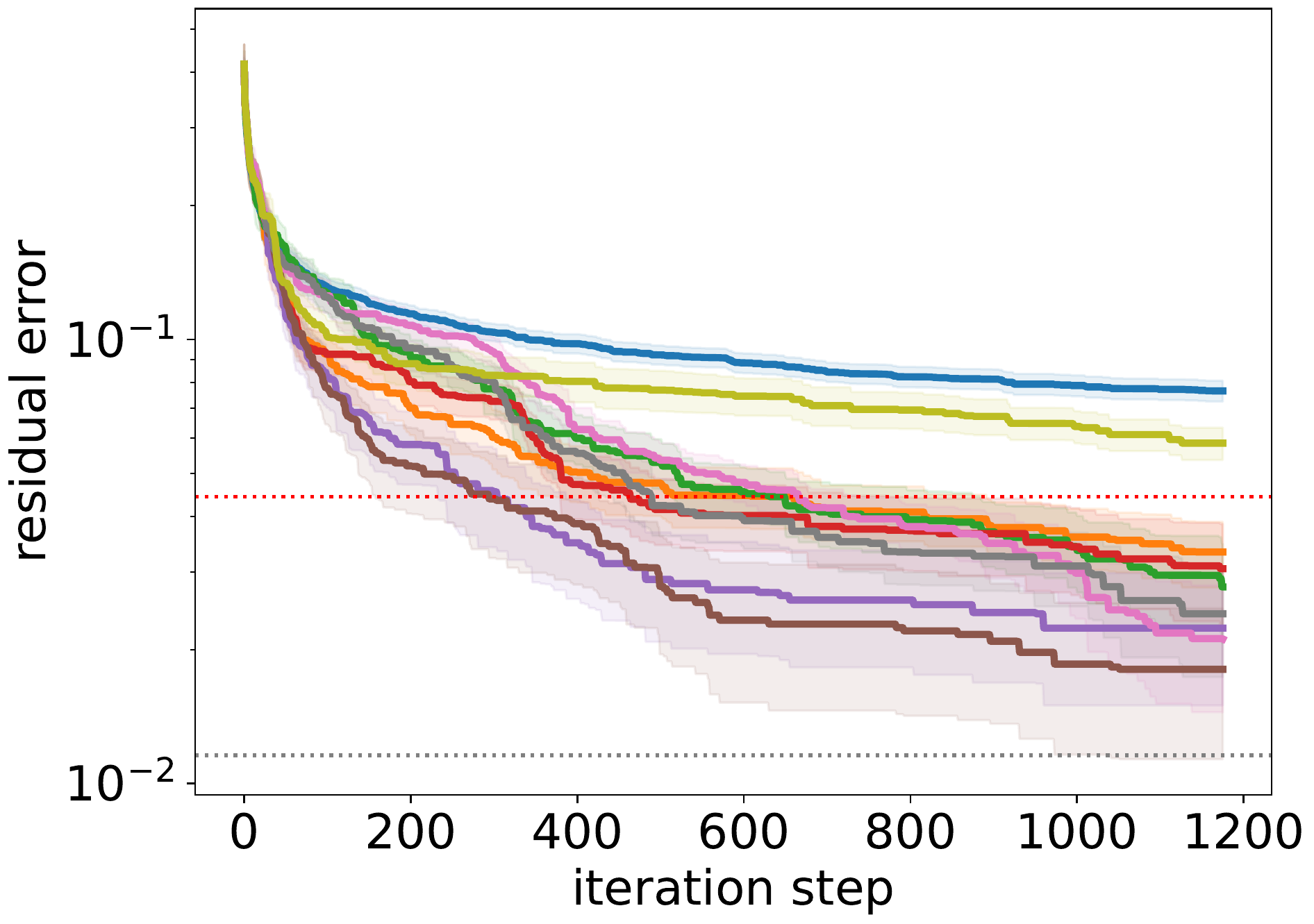}
    \includegraphics[width=50mm]{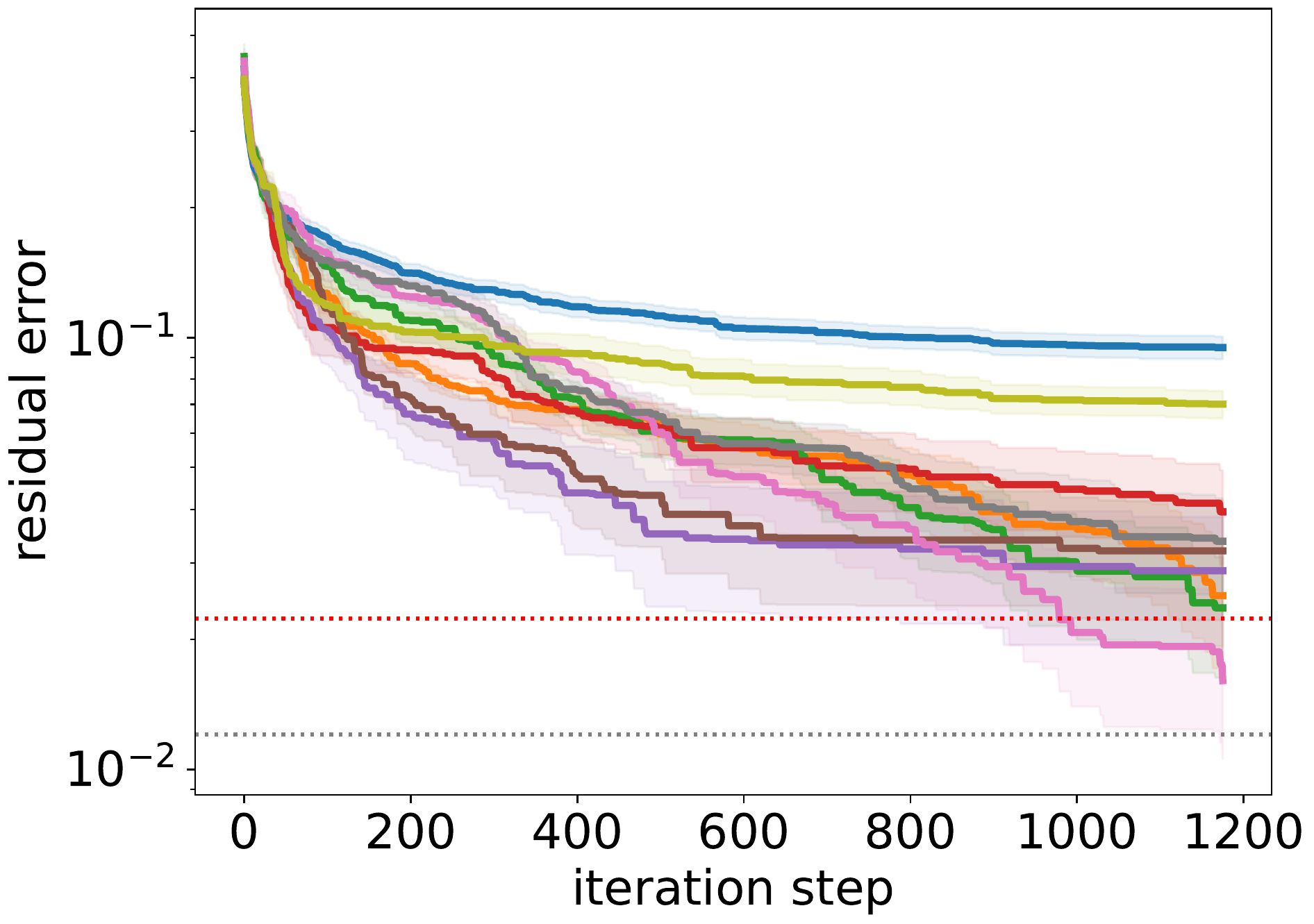}
    \includegraphics[width=50mm]{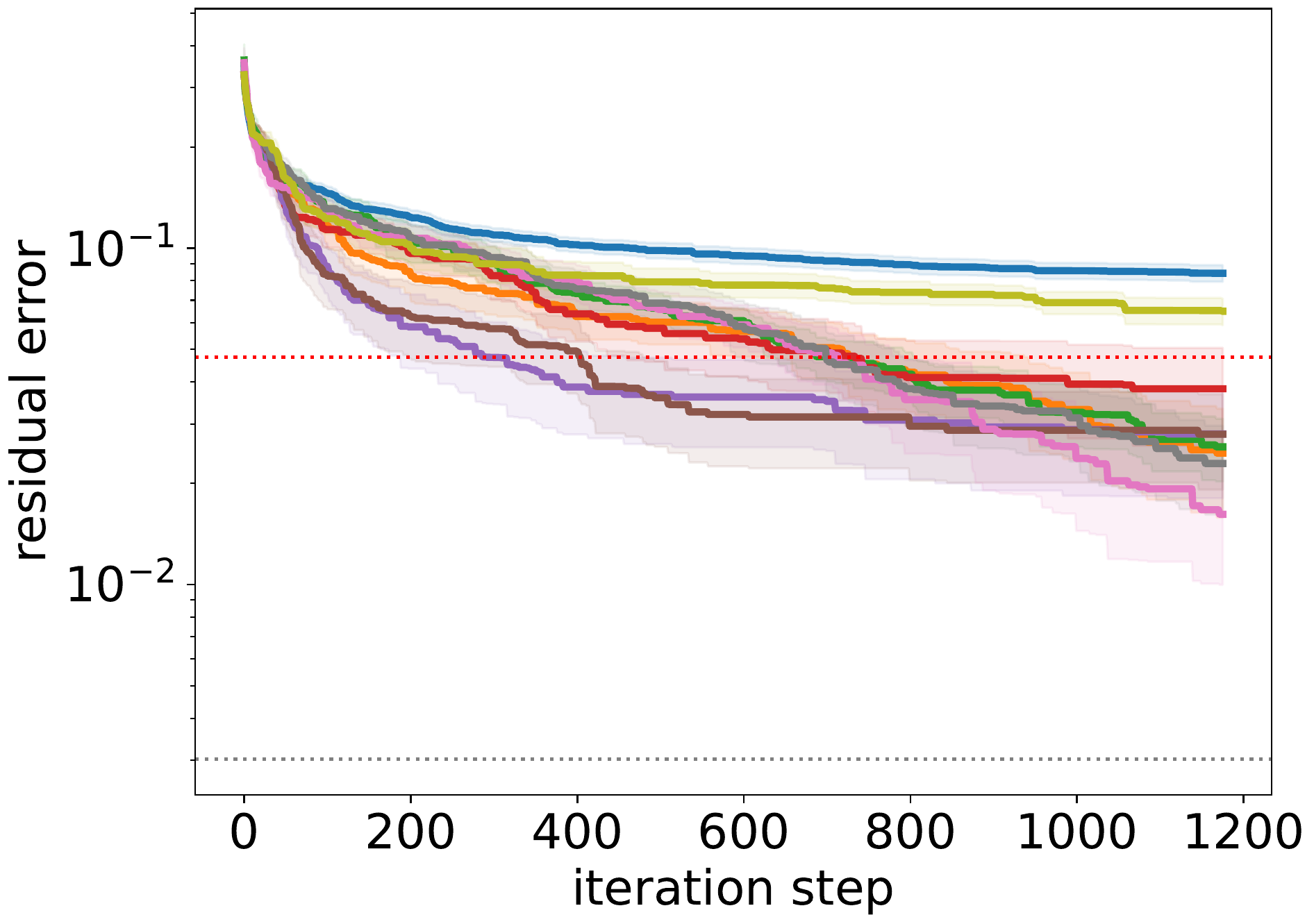}
    \includegraphics[width=50mm]{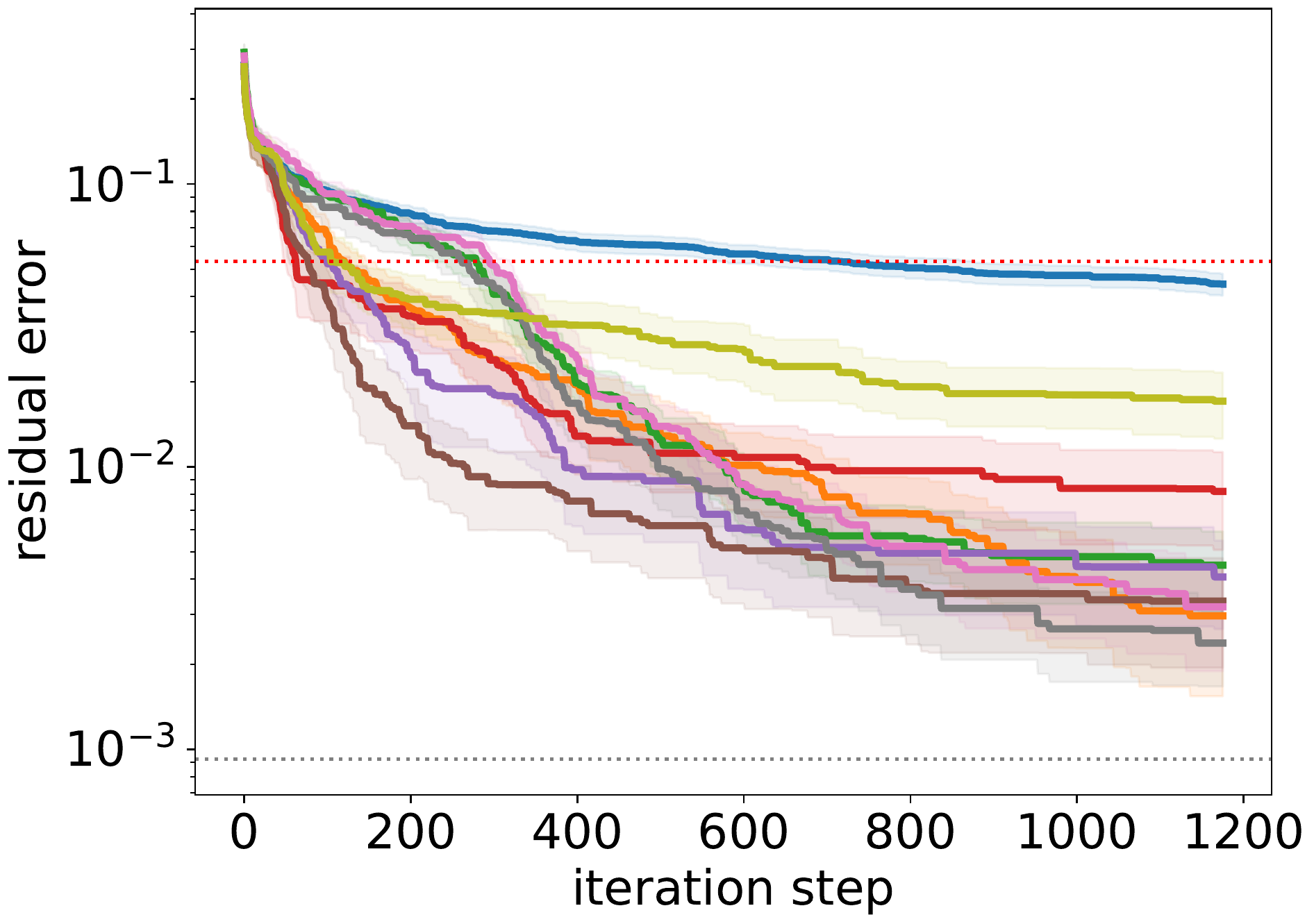}
    \caption{
        \label{fig_s3}
        The residual error of lossy compression of the other nine matrices (instances) as a function of the iteration step among various algorithms with 95\% confidence intervals.
        The errors of the nine exact solutions (baseline of the plots) are 0.535, 0.388, 0.509, 0.487, 0.488, 0.379, 0.367, 0.422, and 0.520, respectively.
	}
\end{figure}

Figure~\ref{fig_s3} shows the results of the other 9 instances used in Table~\ref{table1}.

\section*{Data availability}

The program code to reproduce the analyses is in Supplementary Information.


\section*{Acknowledgement}

T.K. thanks Ikuro Sato, Kentaro Matsuura and Takashi Imoto for useful discussions.

\section*{Author contributions statement}

T.K. and M.A. conceived the concept. T.K. conducted the experiments and analysed the results. All authors reviewed the manuscript.

\section*{Competing interests}

The authors declare no competing interests.

\end{document}